\documentclass{article}

\usepackage[final]{conf}

\usepackage[utf8]{inputenc} 
\usepackage[T1]{fontenc}    
\usepackage{url}            
\usepackage{booktabs}       
\usepackage{amsfonts}       
\usepackage{nicefrac}       
\usepackage{microtype}      
\usepackage{lipsum}
\usepackage{fancyhdr}       
\usepackage{graphicx}       
\usepackage{amsmath}
\usepackage{amssymb}
\usepackage{mathtools}
\usepackage{amsthm}
\usepackage{color, soul, colortbl}
\usepackage{multirow}
\usepackage{changepage}

\graphicspath{{media/}}     
\usepackage{fontawesome5}

\usepackage[pagebackref,breaklinks,colorlinks]{hyperref}

\definecolor{myLinkColor}{rgb}{0.18,0.39,0.62}
\hypersetup{
    colorlinks=true,
    linkcolor=myLinkColor,
    filecolor=myLinkColor,
    urlcolor=myLinkColor,
    citecolor=myLinkColor,
}

\makeatletter
\newcommand{\github}[1]{%
   \href{#1}{\faGithubSquare}%
}
\makeatother

\title{h2oGPT: Democratizing Large Language Models
}

\author{
  Arno Candel, Jon McKinney, Philipp Singer, Pascal Pfeiffer, Maximilian Jeblick, \\ Prithvi Prabhu, Jeff Gambera, Mark Landry, Shivam Bansal, Ryan Chesler, Chun Ming Lee, \\ Marcos V. Conde, Pasha Stetsenko, Olivier Grellier, SriSatish Ambati
  \thanks{Please cite this work as ``h2oGPT by H2O.ai". This is work in progress. Correspondence regarding this technical report can be sent to \texttt{\{arno, jon.mckinney, sri\}@h2o.ai}}\\
  \\
  \textbf{H2O.ai, Inc.}\\
  Mountain View, CA\\
}

\begin{document}

\maketitle

\begin{figure}[h]
    \centering
    \vspace{-18pt}
    \includegraphics[width=6cm]{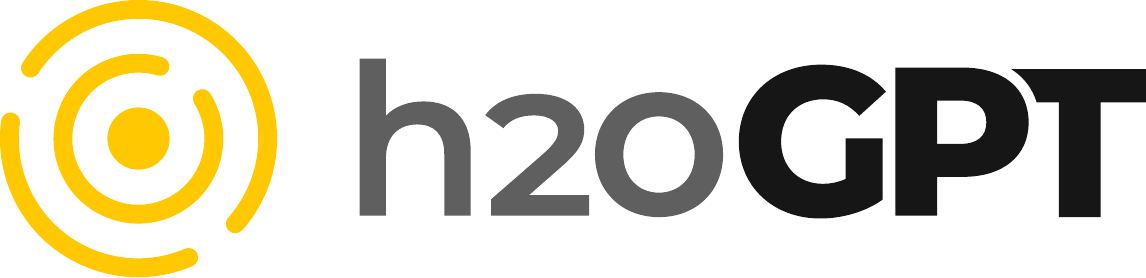}\\
    \vspace{2mm}
    {\large\url{https://github.com/h2oai/h2ogpt}}\\
    {\large\url{https://gpt.h2o.ai}}\\
    \vspace{10pt}
\end{figure}

\begin{figure}[h]
    \centering
    \vspace{-10pt}
    \includegraphics[width=6cm]{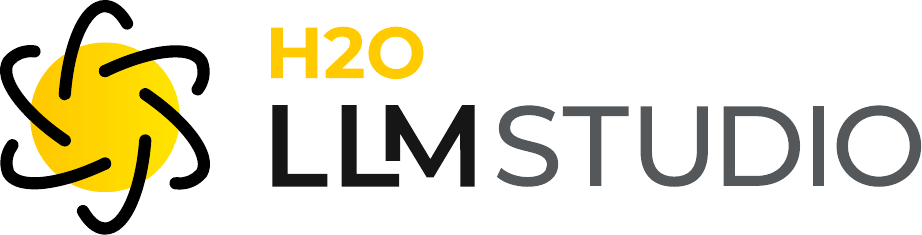}\\
    \vspace{2mm}
    {\large\url{https://github.com/h2oai/h2o-llmstudio}}\\
    \vspace{15pt}
\end{figure}

\begin{abstract}
Applications built on top of Large Language Models (LLMs) such as GPT-4 represent a revolution in AI due to their human-level capabilities in natural language processing. However, they also pose many significant risks such as the presence of biased, private, or harmful text, and the unauthorized inclusion of copyrighted material.

We introduce h2oGPT, a suite of open-source code repositories for the creation and use of LLMs based on Generative Pretrained Transformers (GPTs). The goal of this project is to create the world's best truly open-source alternative to closed-source approaches. In collaboration with and as part of the incredible and unstoppable open-source community, we open-source several fine-tuned h2oGPT models from 7 to 40 Billion parameters, ready for commercial use under fully permissive Apache 2.0 licenses. Included in our release is 100\% private document search using natural language.

Open-source language models help boost AI development and make it more accessible and trustworthy. They lower entry hurdles, allowing people and groups to tailor these models to their needs. This openness increases innovation, transparency, and fairness. An open-source strategy is needed to share AI benefits fairly, and H2O.ai will continue to democratize AI and LLMs.

\end{abstract}

\begin{adjustwidth}{37pt}{37pt}
\end{adjustwidth}

\clearpage

\tableofcontents

\vspace{8mm}
\section*{Transparency and Accessibility}
This is an open-source project, the code and models are publicaly available, free of charge. 

The official GitHub repository for h2oGPT is \url{https://github.com/h2oai/h2ogpt}, and for LLM Studio is \url{https://github.com/h2oai/h2o-llmstudio}, both are open to contributions from the community and in constant evolution.

The foundation large language models (LLMs) presented in this work, can be tested in our online playground \url{https://gpt.h2o.ai/} --- no login required, completely free.

\clearpage

\section{Introduction}
Recent advances in LLMs and GPTs are all over the news. Companies like OpenAI, Google, Anthropic, Microsoft, Cohere, Meta, Stability.AI, AI21 Labs, and many others have established leadership in the development and integration of LLMs. However, none of the above companies are providing truly open-source commercially viable models or even training data.


H2O.ai has built several world-class Machine Learning, Deep Learning and AI platforms over the past decade, much of it as open-source software (and on top of existing open-source software), and has earned the trust of its customers across the globe. We are ideally positioned to provide an open-source GPT ecosystem to enterprises, organizations, and individuals across the world.

\subsection{Why Open-Source LLMs?}
Every nation, state, and city needs its own GPT. This is because LLMs can be used for a variety of purposes, such as health care, science, and education.

While commercially hosted and centralized LLMs like OpenAI's ChatGPT/GPT-4, Anthropic's Claude, Microsoft's Bing AI Chat, Google's Bard, and Cohere are powerful and effective, they have certain limitations compared to open-source LLMs:
\begin{itemize}
    \item \textbf{Data Privacy and Security}: Using hosted LLMs requires sending data to external servers. This can raise concerns about data privacy, security, and compliance, especially for sensitive information or industries with strict regulations.
    \item \textbf{Dependency and Customization}: Hosted LLMs often limit the extent of customization and control, as users rely on the service provider's infrastructure and predefined models. Open-source LLMs allow users to tailor the models to their specific needs, deploy on their own infrastructure, and even modify the underlying code.
    \item \textbf{Cost and Scalability}: Hosted LLMs usually come with usage fees, which can increase significantly with large-scale applications. Open-source LLMs can be more cost-effective, as users can scale the models on their own infrastructure without incurring additional costs from the service provider.
    \item  \textbf{Access and Availability}: Hosted LLMs may be subject to downtime or limited availability, affecting users' access to the models. Open-source LLMs can be deployed on-premises or on private clouds, ensuring uninterrupted access and reducing reliance on external providers.
\end{itemize}
Overall, open-source LLMs offer greater flexibility, control, and cost-effectiveness, while addressing data privacy and security concerns. They foster a competitive landscape in the AI industry and empower users to innovate and customize models to suit their specific needs.

\section{The Making of h2oGPT}
In this section, we detail some of the work done to create the fine-tuned h2oGPT models we released. We show what data and models were used in the process.  More detail can be found on \href{https://github.com/h2oai/h2ogpt/issues}{\faGithubSquare h2oGPT GitHub issues} and \href{https://github.com/h2oai/h2o-llmstudio/issues}{\faGithubSquare H2O LLM Studio GitHub issues}.
\label{sec:headings}
\subsection{Foundation Models and Datasets}
To create a conversational GPT, we need a foundation model that can generate tokens, and we need to fine-tune it to become conversational (i.e., create useful answers for given prompts). One can also fine-tune a foundation model to become good at summarizing articles, or good at converting articles into JSON key/value pairs etc., but the key is a good foundation model and a small but high-quality dataset for fine-tuning.

\subsubsection{Pre-Training vs Fine-Tuning}
\begin{itemize}
    \item \textbf{Pre-training}: Typically on TBs of data, gives the LLM the ability to master one or many \textbf{languages}. Pre-training usually takes weeks or months on dozens or hundreds of GPUs. The most common concern is underfitting and cost.
    \item \textbf{Fine-tuning}: Typically on MBs or GBs of data, makes a model more familiar with a \textbf{specific style} of prompting, which generally leads to improved outcomes for this one specific case. The most common concern is overfitting. Fine-tuning usually takes hours or days on a few GPUs. 
\end{itemize}

\subsubsection{Foundation Models}
The following permissively licensed foundation models are available currently (May 2023), in Hugging Face format, for easy adoption:
\begin{itemize}
    \item EleutherAI/pythia-6.9b
    \item EleutherAI/pythia-12b and EleutherAI/pythia-12b-deduped
    \item \textbf{EleutherAI/gpt-neox-20b}
    \item mosaicml/mpt-7b-storywriter
    \item tiiuae/falcon-7b
    \item \textbf{ttiuae/falcon-40b}
    \item bigscience/bloom
\end{itemize}
The largest foundation models we used were \href{https://arxiv.org/abs/2204.06745}{GPT-NeoX-20B: An Open-Source Autoregressive Language Model} (from April 2022), and \href{https://huggingface.co/tiiuae/falcon-40b}{Falcon-40B} (from May 2023). The largest available fully open-source model to this day is \href{https://huggingface.co/bigscience/bloom}{Bloom 176B}, but it is too big to be practical, and also undertrained.
The above models from EleutherAI and bigscience were trained on a relatively small number of tokens using \href{https://arxiv.org/abs/2203.15556}{Chinchilla} scaling laws, but it later turned out that smaller models trained on more tokens can perform even better, such as \href{https://arxiv.org/abs/2302.13971}{LLaMa}, and now Falcon. The above models (except for mpt-7b-storywriter) also have relatively short context lengths of only 2048 tokens (can only summarize about one page), and models with larger context lengths would be preferable for many downstream tasks.

\begin{table*}[t!]
    \center
    \begin{tabular}{lrcccccc}
        \toprule
         &  & Humanities & STEM & Social Sciences & Other & Average\\
        \midrule
        GPT-NeoX (\textbf{h2oGPT})   & 20B   & 29.8 & 34.9 & 33.7 & 37.7 & 33.6 \\
        Falcon (\textbf{h2oGPT}) & 40B & & & & &  54.2 \\
        GPT-3      & 175B  & 40.8 & 36.7 & 50.4 & 48.8 & 43.9 \\
        GPT-4      & ? & & & & & \textbf{86.4} \\
        Gopher     & 280B  & 56.2 & 47.4 & 71.9 & 66.1 & 60.0 \\
        Chinchilla & 70B   & 63.6 & 54.9 & 79.3 & \textbf{73.9} & 67.5\\
        \midrule
        {PaLM}
                   & 8B       & 25.6 & 23.8 & 24.1 & 27.8 & 25.4 \\
                   & 62B      & 59.5 & 41.9 & 62.7 & 55.8 & 53.7 \\
                   & 540B     & \textbf{77.0} & \textbf{55.6 }&\textbf{ 81.0} & 69.6 & 69.3\\
        \midrule
        {LLaMa}
                   & 7B & 34.0 & 30.5 & 38.3 & 38.1 & 35.1 \\
                   & 13B  & 45.0 & 35.8 & 53.8 & 53.3 & 46.9 \\
                   & 33B  & 55.8 & 46.0 & 66.7 & 63.4 & 57.8 \\
                   & 65B  & 61.8 & 51.7 & 72.9 & 67.4 & 63.4  \\
        \bottomrule
    \end{tabular}
    \caption{
    \textbf{Massive Multitask Language Understanding (MMLU).} Five-shot accuracy.
    From \href{https://arxiv.org/abs/2302.13971}{LLaMa paper}. Falcon value from \href{https://github.com/h2oai/h2ogpt/issues/251}{h2oGPT repository}. GPT-4 value from \href{https://arxiv.org/abs/2303.08774}{GPT-4 TR}.
    \label{tab:mmlu}
    }
\end{table*}

Table~\ref{tab:mmlu} shows the placement of h2oGPT in the ecosystem of non-open-source models.

Several efforts by the open-source community are underway to train improved fully open-source permissive (Apache 2.0 license or similar) foundation models:

 \begin{itemize}
     \item \href{https://github.com/openlm-research/open_llama}{Open LLaMa}
     \item \href{https://www.together.xyz/blog/redpajama}{Red Pajama}
     \item \href{https://www.mosaicml.com/blog/mpt-7b}{MosaicML MPT-7B}
 \end{itemize}
We are not currently training our own foundation models, as more community-driven architectural improvements are likely to arrive soon to further improve the performance of the models. Every small architectural change will require training from scratch.

\subsubsection{Foundation Datasets}

All above models (except for Falcon models) were trained on \href{https://www.arxiv-vanity.com/papers/2101.00027/}{the Pile dataset}, 825 GiB of data. This dataset contains some questionable content, as it was sourced from the internet, but the data preparation methods and the dataset \href{https://github.com/EleutherAI/the-pile}{are publicly available}. Falcon models were trained on the \href{https://arxiv.org/pdf/2306.01116.pdf}{RefinedWeb dataset}, which is 2.8 TiB of internet data prepared with enhanced filtering and deduplication methods.

Several efforts are underway to improve the training data for future foundation models:
\begin{itemize}
    \item \href{https://huggingface.co/datasets/CarperAI/pilev2-dev}{Pile V2}
    \item \href{https://www.together.xyz/blog/redpajama}{Red Pajama}
\end{itemize}

\subsection{Fine-Tuning}
Given a suitable foundation model (currently with 7, 12, 20 or 40 billion parameters), we need a fine-tuning dataset and a Linux box with suitable GPUs. \href{https://github.com/h2oai/h2ogpt/blob/main/FINETUNE.md}{More information about fine-tuning is on our GitHub pages}.

\subsubsection{Fine-Tuning Data Preparation}
To fine-tune a model, we typically need to provide both input (i.e., a prompt) and output (i.e., the response). As most data is not in this format, some effort is required to create these pairs. One example of this:
\begin{verbatim}   
{
  "input": "Who are you?",
  "output": "I am h2oGPT.",
}
\end{verbatim}
Obviously, more sophisticated inputs and outputs will lead the model to become more sophisticated, to the degree that the foundation model's abilities allow. If all outputs in the fine-tuning dataset are very brief, then the model will learn to be brief in general. Note that the model will not be able to memorize all the fine-tuning input/output pairs, but mostly learn the \textbf{style} of the fine-tuning data.

In addition to creating input/output pairs, several other things have to be taken care of:
\begin{itemize}
\item Prompt engineering (e.,g., injection of \texttt{<human>:} and \texttt{<bot>:} into the text to indicate input/outputs)
\item data pre-processing (e.g., removing incomplete sentences, too long dialogues)
\item data filtering (e.g., removing profanity or undesired responses that are too short or low quality)
\end{itemize}
So far, we have created over 1800 lines of code for data processing and cleaning, required for efficient fine-tuning.
All the code is available at \href{https://github.com/h2oai/h2ogpt/blob/main/create_data.py}{our GitHub repository}.

Most of our h2oGPT models have been trained on at least a significant part of the recently open-sourced \href{https://huggingface.co/datasets/OpenAssistant/oasst1}{Open Assistant dataset}. It contains high-quality conversational data, sourced from crowd-sourced humans. While the quality isn't as high as e.g., GPT-4 output (which we can't use due to the terms of service), it's the best open-source instruction data available at the moment. The challenge lies in identifying low-quality data points, or in the generation of even higher-quality data points from it, see our roadmap items below.

We use the \href{https://huggingface.co/OpenAssistant/reward-model-deberta-v3-large-v2}{Open Assistant Reward Model} to assign scores to each conversation, and we drop those with low scores.

For fine-tuning our models, we end up with high-quality conversations like these (shown here in fully formatted form for actual training, with \texttt{"<human>: "} and \texttt{"<bot>: "} prefixes to indicate inputs and outputs).

\begin{verbatim}
"<human>: What is the historical origin of werewolf stories? <bot>: The historic 
starting place of werewolf tales may be traced again to historic civilizations, 
consisting of the Greeks and Romans, who informed stories of people who converted 
into wolves. Werewolf legend additionally has roots in European folklore, in which 
it became believed that people may want to change into wolves via a curse or via 
way of means of being bitten via way of means of any other werewolf. The notion 
of werewolves continued via the Middle Ages and have become a famous problem in 
Gothic literature at some point of the Renaissance. <human>: Why is it said that 
silver is needed to kill a werewolf? <bot>: Now, silver sulfide is not shown to 
be toxic to any other animals, but it's not soluble in water, and so can only be
ingested. In a werewolf, it would lift from the silver and travel through the 
bloodstream of the animal, blocking off blood vessels and poisoning cells. And
it is this that kills the werewolf. <human>:"	
\end{verbatim}

The \texttt{"<human>:"} at the end indicates that the response started by the last \texttt{"<bot>:"} is complete. This way, the model learns to properly terminate its responses. Note that the choice of chat-indicators like \texttt{"<human>:"} is arbitrary, and dedicated tokens are often used to avoid tokenization ambiguities.

Training on realistic multi-turn conversations like the one shown above substantially improves generation quality for a chatbot, according to common sense and other research such as \href{https://arxiv.org/abs/2305.11206}{LIMA}. During a chat conversation, prior dialogues are added as part of the context at generation time, so follow-up prompts like \texttt{Summarize the above} are perfectly fine.

We make this dataset available on \href{https://huggingface.co/datasets/h2oai/openassistant_oasst1_h2ogpt_graded}{our Hugging Face page}. It is intended to be used in combination with the fine-tuning methods provided by the \href{https://github.com/h2oai/h2ogpt/blob/main/FINETUNE.md}{h2oGPT repository}.

\subsubsection{H2O LLM Data Studio}
We also improved the foundational scripts used in the data preparation for the h2oGPT model. We added more generalization in the code, comprehensive error handling, handling a variety of training/tuning tasks, and a variety of text cleaning and data preparation utility functions. This led to the development of H2O LLM Data Studio - a toolkit for data preparation for LLM fine-tuning. 

LLM Data Studio can be used to prepare datasets for a variety of downstream tasks, This includes:

\begin{itemize}
    \item \textbf{Question Answering:} It involves preparing datasets that consist of contextual information, questions, and corresponding answers. This task is essential for training question-answering models that can accurately respond to queries based on the provided context. The dataset preparation process focuses on building a well-structured dataset for training question-answering systems.

    \item \textbf{Text Summarization:} It involves preparing datasets that consist of articles and their corresponding summaries. In this task, the dataset preparation process focuses on extracting important information from the articles and creating concise summaries that capture the key points. With the prepared datasets, users can train text summarization models to generate concise and informative summaries from longer pieces of text.

    \item \textbf{Instruct Tuning: }It involves preparing datasets that consist of prompts or instructions and their corresponding responses. This task is essential for training models that effectively understand and adhere to the provided instructions and accurately respond to user prompts.

    \item \textbf{Human-Bot Conversations:} It involves preparing datasets that contain multiple conversations between human users and chat bots. This task is essential for training models that can understand user intents, and provide accurate responses, leading to improved conversational experiences. During dataset preparation, the focus is on structuring and organizing the conversational data, including user queries, bot responses, and any relevant context.

    \item \textbf{Continued Pre-Training:} It involves preparing datasets with long texts to facilitate further pre-training of language models. In this task, the dataset preparation process focuses on organizing long textual data to allow the language models to learn from extensive and diverse linguistic patterns, leading to enhanced language understanding and generation capabilities.
\end{itemize}

Key techniques supported in LLM Data Studio:

\begin{itemize}
    \item Data Augmentation: Augment or mix multiple data sets as a single data object
    \item Text Cleaning: Clean the text using different cleaning methods such as stop words removal, punctuation removal, special character removal, case handling
    \item Profanity Check: Check and remove any texts objects having profanity
    \item Text Quality Check: Check and remove any texts having profanity
    \item Truncate by Length: Truncate the sentence based on a max length parameter
    \item Valid Q\&A: Calculate the similarity score and filter the dataset based on a similarity threshold
    \item Pad Sequence: Pad the sequence based on a maximum length parameter
    \item Truncate Sequence by Score: Truncate the sequence based on a score and max length parameter required for the model.
    \item Output Conversion: Convert the transformed dataset to an output object such as JSON
    \item Compression Ratio Filter: Filter the text summarizing by comparing the compression ratio of the summaries
    \item Boundary Marking: Add start and end tokens in the boundaries of the summary text
\end{itemize}

The typical workflow for data preparation in H2O LLM Studio involves several sequential steps. Firstly, the user performs data ingestion, where they import various types of documents from different connectors. Once the data is ingested, the next step is to select the target training task, which can include tasks like continued pretraining, instruct tuning, chatbot development, or RLHF protection.

After selecting the training task, users have the option to augment their dataset by incorporating additional data from other sources. This data mix-in or augmentation step allows for the enrichment of the existing dataset. 

Subsequently, the data cleaning process takes place, wherein low-quality parts of the data are removed. This includes eliminating problematic elements like long lines of pure spaces or unusual characters that may hinder analysis or modeling.

To ensure data quality, a data quality checking step is implemented. This involves employing techniques like bleu/meteor/similarity or RLHF reward models to identify and filter out data with poor quality. Additional filters, such as length-based filtering (e.g., short concise answers vs. long answers), and checks for profanity can also be applied during this stage.

Once the text has been cleaned and verified for quality, the user selects the target tool for data transformation. This step involves converting the data, along with its associated metadata, into a suitable format such as JSON for utilization in LLM Studio, h2oGPT, or any other target tool.

Lastly, the data is prepared for the target model. Different models may have specific requirements for context length or cutoff length, and the data needs to be adjusted accordingly. This ensures that the text is appropriately truncated to match the desired specifications of the target model, avoiding any truncation issues or poor data representation.

By following this systematic workflow, users can effectively prepare their data for analysis and modeling in H2O LLM Studio, facilitating accurate and reliable research outcomes.

H2O LLM Data Studio is also part of the H2O LLM Ecosystem and is made available to users for the purpose of data cleaning and preparation for fine-tuning LLMs. 

\subsubsection{Fine-Tuning Methods}

\paragraph{LoRA}
We use Huggingface PEFT and its implementation of LoRA (Low Rank Approximation) \href{https://arxiv.org/abs/2106.09685}{LoRA}. This results in substantial speed-up and lower memory use compared to full fine-tuning. Only as a small fraction of weights are trainable, and the required optimizer state is of the order of 20MB instead of 20GB, reducing the memory footprint by at least a factor of 2, and leading to measurable speedups as fewer GPUs are needed and fewer gradients need to be computed. In addition, full fine-tuning can result in catastrophic forgetfulness,
which can be prevented using adapter methods like LoRA by focusing the fine-tuning on specific parts
of the neural network architecture, such as the attention heads.

Injecting LoRA into linear layers turns the dense matrices into read-only weights, and adds a product of two small trainable matrices with a scaling factor, for reduced memory overhead during back-propagation (training).

Original model architecture for the \texttt{h2oai/h2ogpt-oasst1-falcon-40b} model:
\begin{small}
\begin{verbatim}
RWForCausalLM(
  (transformer): RWModel(
    (word_embeddings): Embedding(65024, 8192)
    (h): ModuleList(
      (0-59): 60 x DecoderLayer(
        (ln_attn): LayerNorm((8192,), eps=1e-05, elementwise_affine=True)
        (ln_mlp): LayerNorm((8192,), eps=1e-05, elementwise_affine=True)
        (self_attention): Attention(
          (maybe_rotary): RotaryEmbedding()
          (query_key_value): Linear(in_features=8192, out_features=9216, bias=False)
          (dense): Linear(in_features=8192, out_features=8192, bias=False)
          (attention_dropout): Dropout(p=0.0, inplace=False)
        )
        (mlp): MLP(
          (dense_h_to_4h): Linear(in_features=8192, out_features=32768, bias=False)
          (act): GELU(approximate='none')
          (dense_4h_to_h): Linear(in_features=32768, out_features=8192, bias=False)
        )
      )
    )
    (ln_f): LayerNorm((8192,), eps=1e-05, elementwise_affine=True)
  )
  (lm_head): Linear(in_features=8192, out_features=65024, bias=False)
)
\end{verbatim}
\end{small}
After adding LoRA adapters for the \texttt{Linear} layers (dense matrix multiplies), we get the following model architecture for the trainable weights:
\begin{small}
\begin{verbatim}
PeftModelForCausalLM(
  (base_model): LoraModel(
    (model): RWForCausalLM(
      (transformer): RWModel(
        (word_embeddings): Embedding(65024, 8192)
        (h): ModuleList(
          (0-59): 60 x DecoderLayer(
            (ln_attn): LayerNorm((8192,), eps=1e-05, elementwise_affine=True)
            (ln_mlp): LayerNorm((8192,), eps=1e-05, elementwise_affine=True)
            (self_attention): Attention(
              (maybe_rotary): RotaryEmbedding()
              (query_key_value): Linear8bitLt(
                in_features=8192, out_features=9216, bias=False
                (lora_dropout): ModuleDict(
                  (default): Dropout(p=0.05, inplace=False)
                )
                (lora_A): ModuleDict(
                  (default): Linear(in_features=8192, out_features=8, bias=False)
                )
                (lora_B): ModuleDict(
                  (default): Linear(in_features=8, out_features=9216, bias=False)
                )
                (lora_embedding_A): ParameterDict()
                (lora_embedding_B): ParameterDict()
              )
              (dense): Linear8bitLt(
                in_features=8192, out_features=8192, bias=False
                (lora_dropout): ModuleDict(
                  (default): Dropout(p=0.05, inplace=False)
                )
                (lora_A): ModuleDict(
                  (default): Linear(in_features=8192, out_features=8, bias=False)
                )
                (lora_B): ModuleDict(
                  (default): Linear(in_features=8, out_features=8192, bias=False)
                )
                (lora_embedding_A): ParameterDict()
                (lora_embedding_B): ParameterDict()
              )
              (attention_dropout): Dropout(p=0.0, inplace=False)
            )
            (mlp): MLP(
              (dense_h_to_4h): Linear8bitLt(
                in_features=8192, out_features=32768, bias=False
                (lora_dropout): ModuleDict(
                  (default): Dropout(p=0.05, inplace=False)
                )
                (lora_A): ModuleDict(
                  (default): Linear(in_features=8192, out_features=8, bias=False)
                )
                (lora_B): ModuleDict(
                  (default): Linear(in_features=8, out_features=32768, bias=False)
                )
                (lora_embedding_A): ParameterDict()
                (lora_embedding_B): ParameterDict()
              )
              (act): GELU(approximate='none')
              (dense_4h_to_h): Linear8bitLt(
                in_features=32768, out_features=8192, bias=False
                (lora_dropout): ModuleDict(
                  (default): Dropout(p=0.05, inplace=False)
                )
                (lora_A): ModuleDict(
                  (default): Linear(in_features=32768, out_features=8, bias=False)
                )
                (lora_B): ModuleDict(
                  (default): Linear(in_features=8, out_features=8192, bias=False)
                )
                (lora_embedding_A): ParameterDict()
                (lora_embedding_B): ParameterDict()
              )
            )
          )
        )
        (ln_f): LayerNorm((8192,), eps=1e-05, elementwise_affine=True)
      )
      (lm_head): Linear(in_features=8192, out_features=65024, bias=False)
    )
  )
)
trainable params: 55541760 || all params: 41358835712 || trainable%: 0.13429236835089367
\end{verbatim}
\end{small}
The resulting number of trainable parameters is typically around 0.1\% of the original weights, and the degree of approximation can be parameterized with several tuning parameters, most of which don't seem to have a large impact on accuracy, which is great. This makes LoRA one of the most useful techniques for efficient fine-tuning.

\paragraph{bitsandbytes}
To further reduce memory requirements on costly GPU hardware, we make use of 16-bit, 8-bit or 4-bit training using mixed precision hardware and software support, instead of 32-bit or 64-bit precision, which are commonly used across most computing applications. The benefit of the speedup and cost savings from being able to fit the entire model into one GPU is much higher than the downside due to loss of precision. Training or inference with the base model in 8-bit or 4-bit is achieved using PEFT and \href{https://github.com/TimDettmers/bitsandbytes}{bitsandbytes}. While this lowers the memory cost by about a factor of two compared to the use of LoRA alone, it is substantially slower for training than 16-bit on current architectures. Training using 4-bit precision was just made possible and should help with further democratizing LLM fine-tuning to consumer GPUs with 24GB of VRAM or less, cf~\href{https://arxiv.org/abs/2305.14314}{QLoRA}.

Native training using 8-bit floating point precision developed by NVIDIA on H100 GPUs should lead to significant memory savings without compromising training speed, but we haven't had a chance to try that yet.

\subsubsection{Fine-Tuning Hardware requirements}
\paragraph{NVIDIA GPUs}
Using LoRA and 8-bit training, we can fine-tune LLMs with 20B parameters on commodity GPUs with 24GB of VRAM, but just barely, and only for short input/outputs (token length), with batch size 1. We recommend A100 or A6000 (Ada) NVIDIA cards for fine-tuning, or H100, to get the best price/performance, or the use of 4-bit training for cards with less VRAM.

These are the minimally recommended GPU memory sizes for fine-tuning the respective h2oGPT models and 16-bit training is recommended wherever possible, as it can be much faster (by a factor 4 over 8-bit, 4-bit performance is not yet widely tested):

\begin{table}[h]
\centering
\begin{tabular}{ c c c c }
\toprule
\textbf{h2oGPT Model Size} & \textbf{4-bit} & \textbf{8-bit} & \textbf{16-bit} \\
\midrule
7B & 16GB & 12GB & 16GB \\

12B & 16GB & 24GB & 32GB\\

20B & 16GB & 32GB & 48GB\\

30B (research) & 24GB & 48GB & 80GB\\

40B & 48GB & 80GB & 2x80GB\\

65B (research) & 48GB & 80GB & 2x80GB\\
\bottomrule
\end{tabular}
\vspace{1mm}
\caption{h2oGPT model size comparison.}
\end{table}

16GB/32GB cards include V100, 24GB cards include 3090/4090, 40GB cards include A100, 48GB cards include A6000/A6000 Ada, 80GB cards include A100/H100.

Training on multiple GPUs is always faster than training on one GPU, and data parallelism is enabled by default. Larger GPU memory sizes can allow faster training too, since more training data can be streamed. For example, if the model requires 20GB of memory, then one 80GB GPU might allow a batch size of 8, while a 24GB card can only fit a batch size of 1. Having 8x80GB can hence lead to a significant speedup compared to 1x24GB etc. Multi-node multi-GPU training is also possible in the existing framework, and LoRA training requires minimal communication between nodes, which makes it feasible to train on nodes with low interconnect speeds.

We did not try fine-tuning with TPUs or other accelerators, as NVIDIA GPUs are currently the best-supported most available hardware.

\section{Results}
Using the methods outlined above, our makers at H2O.ai have created suitable fine-tuning datasets, prompt engineering techniques, fine-tuning methods, UIs, chatbots, and VectorDB-based private document chat systems, and we are open-sourcing everything.

\subsection{The H2O.ai LLM Ecosystem}
Our open-source LLM ecosystem currently includes the following components:
\begin{itemize}
\item \textbf{Code, data, and models}: Fully permissive, commercially usable code, curated fine-tuning data, and fine-tuned models ranging from 7 to 20 billion parameters.
\item \textbf{State-of-the-art fine-tuning}: We provide code for highly efficient fine-tuning, including targeted data preparation, prompt engineering, and computational optimizations to fine-tune LLMs with up to 20 billion parameters (even larger models expected soon) in hours on commodity hardware or enterprise servers. Techniques like low-rank approximations (LoRA) and data compression allow computational savings of several orders of magnitude.
\item \textbf{Chatbot}: We provide code to run a multi-tenant chatbot on GPU servers, with an easily shareable end-point and a Python client API, allowing you to evaluate and compare the performance of fine-tuned LLMs.
\item \textbf{Document Chat using VectorDB}: We provide code for a fully functional natural language-based document search system using Vector databases and prompt engineering. Of course, 100\% private, and no internet connection is needed.
\item \textbf{H2O LLM Studio}: Our no-code LLM fine-tuning framework created by the world's top Kaggle grandmasters makes it even easier to fine-tune and evaluate LLMs. H2O LLM Studio democratizes LLMs for everyone. This means that anyone can use H2O LLM Studio to fine-tune large open-source LLMs like h2oGPT and others on their own private data and on their servers.
\end{itemize}
The links to our open-source repositories and discussion channels are:
\begin{itemize}
\item \href{https://github.com/h2oai/h2ogpt}{\faGithubSquare h2oGPT \texttt{https://github.com/h2oai/h2ogpt}}
\item \href{https://github.com/h2oai/h2o-llmstudio}{\faGithubSquare H2O LLM Studio \texttt{https://github.com/h2oai/h2o-llmstudio}}
\item \href{https://huggingface.co/h2oai}{H2O.ai on Hugging Face \texttt{https://huggingface.co/h2oai}}
\item \href{https://discord.com/channels/1097462770674438174/1100717863221870643}{H2O.ai Generative Discord Channel}
\end{itemize}

Everything we release is based on fully permissive data and models (exceptions such as LLaMa-based models are explicitly marked as research only), with all code open-sourced, enabling broader access for businesses and commercial products without legal concerns, thus expanding access to cutting-edge AI while adhering to licensing requirements.

\subsubsection{h2oGPT models on Hugging Face}
We are making our models available on the \href{https://huggingface.co/h2oai}{Hugging Face} repository.
Notable models include:
\begin{itemize}
    \item \href{https://huggingface.co/h2oai/h2ogpt-oasst1-falcon-40b}{\texttt{h2oai/h2ogpt-oasst1-falcon-40b}}
    \item \href{https://huggingface.co/h2oai/h2ogpt-oig-oasst1-falcon-40b}{\texttt{h2oai/h2ogpt-oig-oasst1-falcon-40b}}
    \item \href{https://huggingface.co/h2oai/h2ogpt-oasst1-512-20b}{\texttt{h2oai/h2ogpt-oasst1-512-20b}}
    \item \href{https://huggingface.co/h2oai/h2ogpt-oasst1-512-12b}{\texttt{h2oai/h2ogpt-oasst1-512-12b}}
    \item \href{https://huggingface.co/h2oai/h2ogpt-oig-oasst1-512-6_9b}{\texttt{h2oai/h2ogpt-oig-oasst1-512-6\_9b}}
    \item \href{https://huggingface.co/h2oai/h2ogpt-gm-oasst1-en-2048-falcon-40b-v1}{\texttt{h2oai/h2ogpt-gm-oasst1-en-2048-falcon-40b-v1}}
    \item \href{https://huggingface.co/h2oai/h2ogpt-gm-oasst1-en-1024-20b}{\texttt{h2oai/h2ogpt-gm-oasst1-en-1024-20b}}
    \item \href{https://huggingface.co/h2oai/h2ogpt-gm-oasst1-en-2048-falcon-7b-v2}{\texttt{h2oai/h2ogpt-gm-oasst1-en-2048-falcon-7b-v2}}
    \item \href{https://huggingface.co/h2oai/h2ogpt-research-oasst1-512-30b}{\texttt{h2oai/h2ogpt-research-oasst1-512-30b}} (non-commercial)
    \item \href{https://huggingface.co/h2oai/h2ogpt-research-oasst1-512-65b}{\texttt{h2oai/h2ogpt-research-oasst1-512-65b}} (non-commercial)
\end{itemize}

To use the models from Python is easy:

\begin{verbatim}
!pip install transformers==4.29.2
!pip install accelerate==0.19.0
!pip install torch==2.0.1
!pip install einops==0.6.1

import torch
from transformers import pipeline, AutoTokenizer
tokenizer = AutoTokenizer.from_pretrained("h2oai/h2ogpt-oasst1-falcon-40b", 
padding_side="left")
generate_text = pipeline(model="h2oai/h2ogpt-oasst1-falcon-40b",
    tokenizer=tokenizer, torch_dtype=torch.bfloat16, trust_remote_code=True, 
    device_map="auto", prompt_type="human_bot")

res = generate_text("Why is drinking water so healthy?", max_new_tokens=100)
print(res[0]["generated_text"])

>>> Drinking water is healthy because it helps to keep your body hydrated and functioning
>>> properly. It also helps to flush out toxins and waste from the body, which can help
>>> to improve your overall health. Additionally, drinking water can help to regulate
>>> your body temperature, which can help to prevent dehydration and heat exhaustion.
\end{verbatim}

\subsubsection{ChatBot}
\href{https://github.com/h2oai/h2ogpt}{\faGithubSquare h2oGPT \texttt{https://github.com/h2oai/h2ogpt}} contains a simple chatbot GUI and client/server API based on \href{https://github.com/gradio-app/gradio}{Gradio}.

\begin{verbatim}
python generate.py --base_model=h2oai/h2ogpt-oasst1-512-12b
\end{verbatim}

\begin{center}
\includegraphics[width=0.8\textwidth]{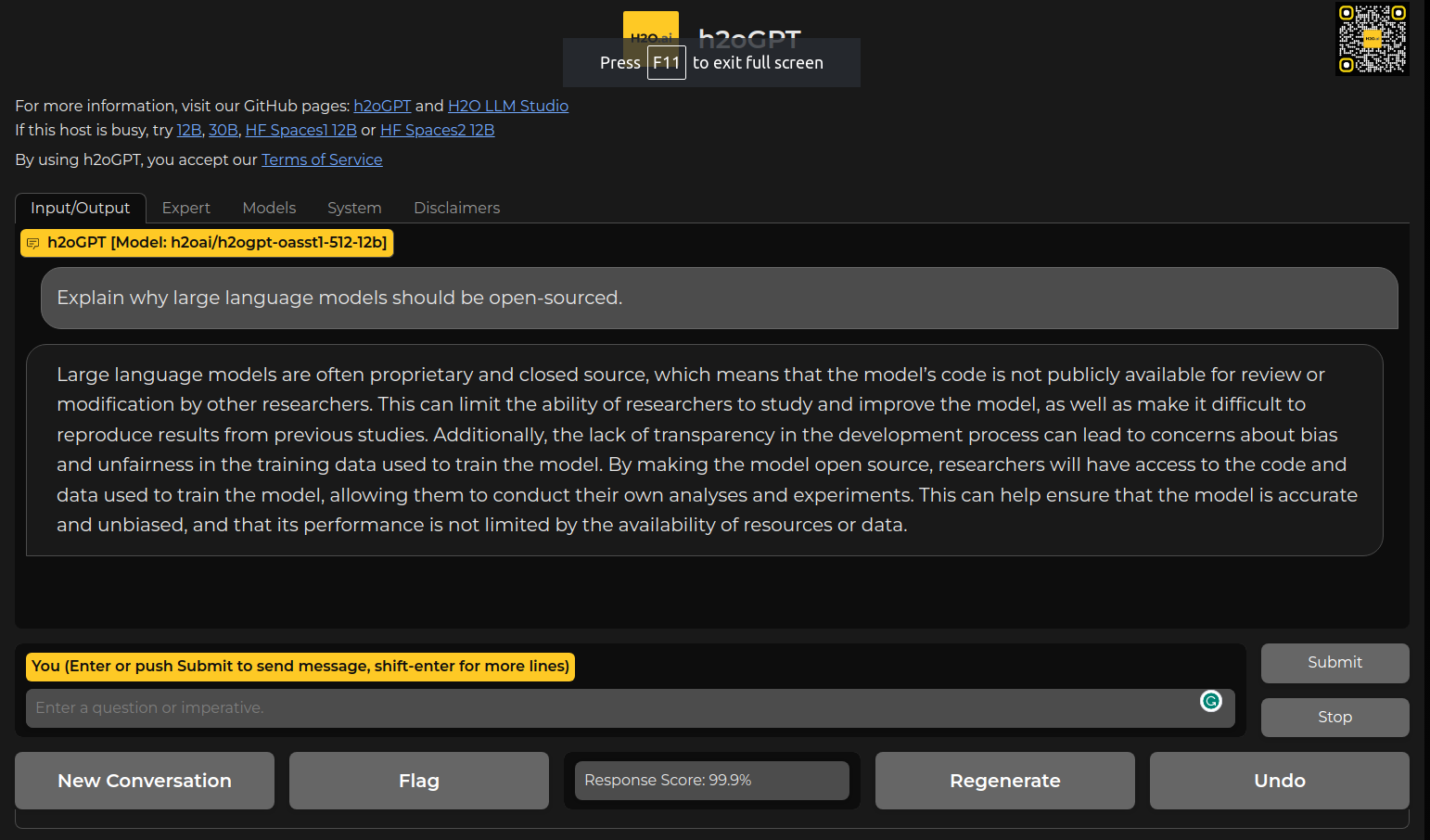}
\end{center}

Chatbot features include:

\begin{itemize}
    \item supports any open-source LLM from Hugging Face
    \item offline mode with no internet access required
    \item comparison of any 2 models
    \item supports LoRA adapter weights on top of any LLM
    \item multi-GPU sharding
    \item automatic scoring of responses using a reward model trained on human feedback
    \item 4-bit quantization options
    \item automatic expansion of context from multiple back-and-forth conversations
\end{itemize}

\subsubsection{Private Document Chat}
It is well-known that LLMs can hallucinate or confabulate their responses, c.f.~\href{https://dl.acm.org/doi/10.1145/3442188.3445922}{On the Dangers of Stochastic Parrots}. It is an active area of research to understand under what conditions this occurs and how to contain it. One way to ground LLMs is to provide source content as context for any query. The query and source content are embedded and similarity is estimated using a vector database. h2oGPT includes FAISS in-memory and Chroma persistent vector databases, relying upon instruct-tuned LLMs to answer the question given the context of top \texttt{k} chunks of source content.

\begin{verbatim}
python generate.py --base_model=h2oai/h2ogpt-research-oasst1-512-30b 
  --langchain_mode=wiki_full
\end{verbatim}

\begin{center}
  \includegraphics[width=0.9\textwidth]{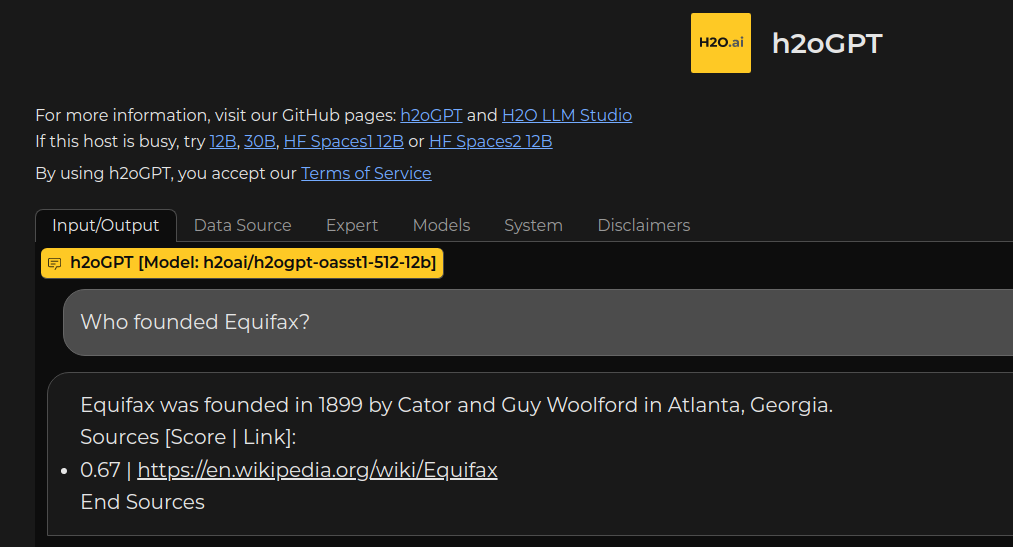}  
\end{center}

Document chat features include:
\begin{itemize}
    \item fact-based question answering for documents
    \item 20GB Wikipedia state is pre-loaded
    \item offline mode with no internet access required
    \item persistent database with vector embeddings
    \item ability to ingest various document types
\end{itemize}

\subsubsection{No-Code Fine-Tuning with H2O LLM Studio}
\href{https://github.com/h2oai/h2o-llmstudio}{\faGithubSquare H2O LLM Studio \texttt{https://github.com/h2oai/h2o-llmstudio}} is an open-source framework that offers both a no-code graphical user interface (GUI) and a command-line interface (CLI) for fine-tuning LLMs. It allows users to train and tweak state-of-the-art LLMs with a variety of hyperparameters, without requiring any coding experience. It supports various advanced finetuning techniques such as Low-Rank Adaptation (LoRA) and 8-bit model training with a low memory footprint. The software allows users to track and compare model performance visually and provides an option to chat with the model for instant performance feedback. Additionally, it facilitates easy model export to the Hugging Face Hub for sharing with the community.

The latest updates to H2O LLM Studio include storing experiment configurations in YAML format and added functionality for supporting nested conversations in data. The system requirements include Ubuntu 16.04+ and an NVIDIA GPU with driver version >= 470.57.02. The software also supports Docker for easy deployment, and it expects CSV input with at least two columns - one for the instruct column and another for the model's expected answer.

Starting H2O LLM Studio is easy:
\begin{verbatim}
make wave
\end{verbatim}

\begin{center}
  \includegraphics[width=0.9\textwidth]{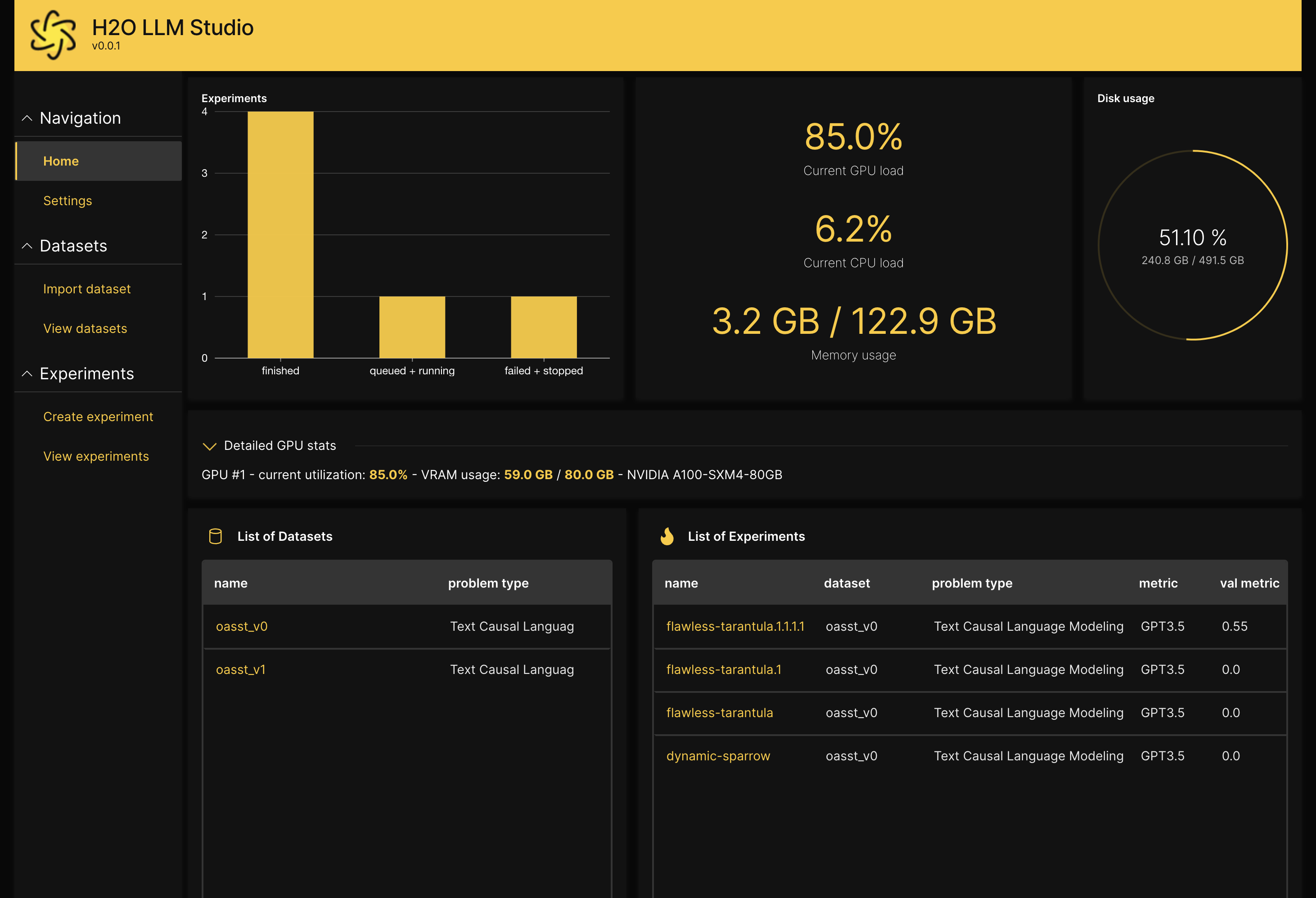}
  
  \includegraphics[width=0.9\textwidth]{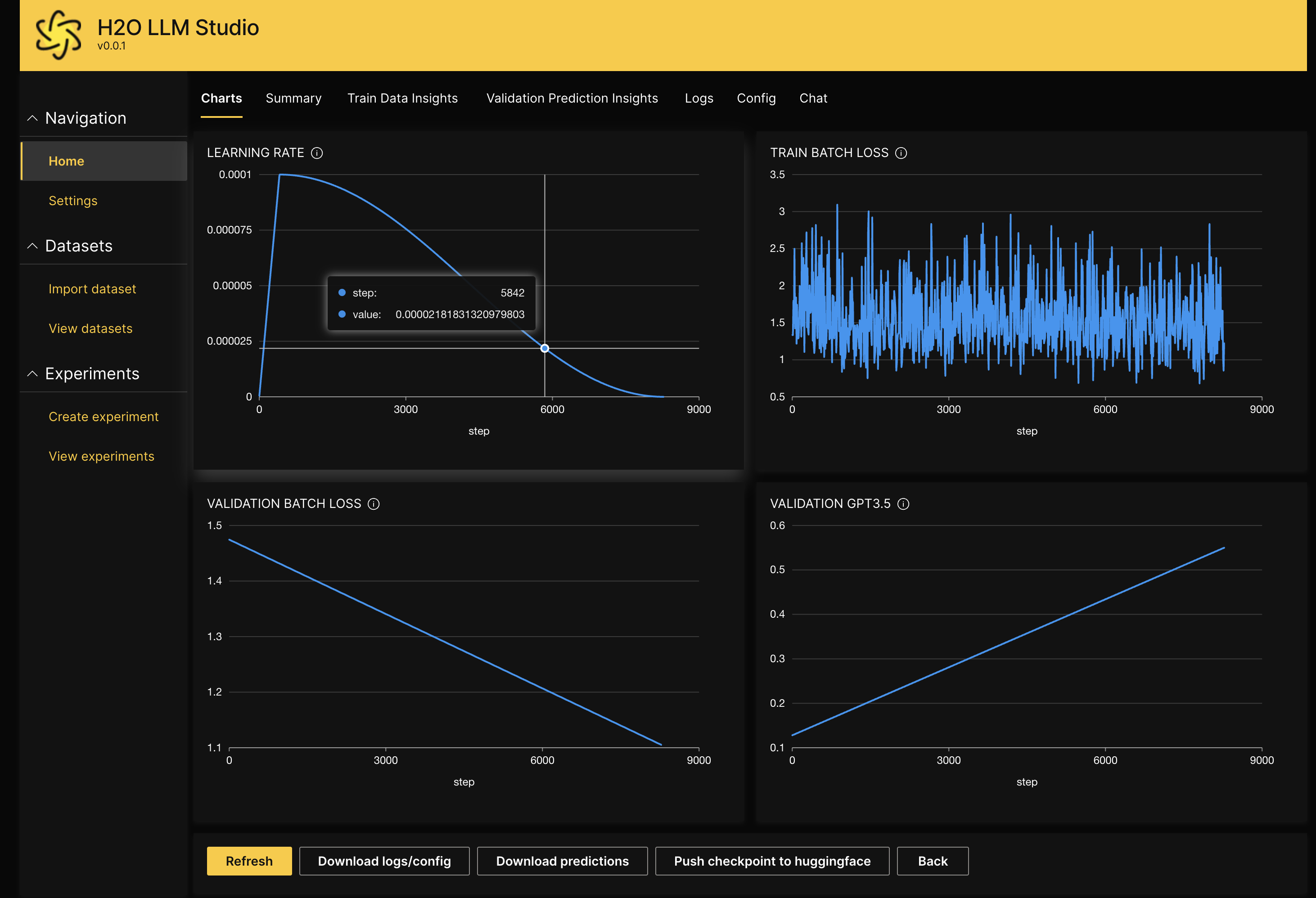}
\end{center}

H2O LLM Studio features include:
\begin{itemize}
    \item easily and effectively fine-tune LLMs without the need for any coding experience
    \item use a graphic user interface (GUI) specially designed for large language models
    finetune any LLM using a large variety of hyperparameters
    \item use recent finetuning techniques such as Low-Rank Adaptation (LoRA) and 8-bit model training with a low memory footprint
    \item use advanced evaluation metrics to judge generated answers by the model
    \item track and compare your model performance visually. In addition, Neptune integration can be used.
    \item chat with your model and get instant feedback on your model performance
    \item easily export your model to the Hugging Face Hub and share it with the community
\end{itemize}

\subsection{Validation, Limitations, and Capabilities}
We are aware that open-source LLMs with fully permissive licenses are not as capable as certain closed-sourced offerings. As the open-source community continues to learn and improve, the available models will become better, and reach a point where they will be more and more suited for commercial applications.

\subsubsection{Evaluation Metrics}
We used the \href{https://github.com/EleutherAI/lm-evaluation-harness}{EleutherAI evaluation harness} to confirm that our fine-tuned LLMs still exhibit the same basic capabilities as the foundation models. Table~\ref{tab:commonsense} shows a comparison of performance on several common-sense reasoning tasks. Note that error bars are on the order of +/- 1.

\begin{table*}[t!]
  \centering
  \setlength{\tabcolsep}{5pt}
  \begin{tabular}{lrccccccccc}
  \toprule
  & & BoolQ & PIQA & \hspace{-0.3cm} HellaSwag \hspace{-0.2cm} & \hspace{-0.2cm} WinoGrande \hspace{-0.3cm} & ARC-e & ARC-c & OBQA \\
  \midrule
  GPT-3        & 175B & 60.5 & 81.0 & 78.9 & 70.2 & 68.8 & 51.4 & 57.6 \\
  Gopher       & 280B & 79.3 & 81.8 & 79.2 & 70.1 & -    & -    & -    \\
  Chinchilla   & 70B  & 83.7 & 81.8 & 80.8 & 74.9 & -    & -    & -    \\
  PaLM         & 62B  & 84.8 & 80.5 & 79.7 & 77.0 & 75.2 & 52.5 & 50.4 \\
  PaLM-cont    & 62B  & 83.9 & 81.4 & 80.6 & 77.0 & -    & -    & -    \\
  PaLM         & 540B & \textbf{88.0} & 82.3 & 83.4 & \textbf{81.1} & 76.6 & 53.0 & 53.4 \\
  \midrule
  {LLaMa}
     & 7B  & 76.5 & 79.8          & 76.1          & 70.1 & 72.8          & 47.6          & 57.2 \\
     & 13B & 78.1 & 80.1          & 79.2          & 73.0 & 74.8          & 52.7          & 56.4 \\
     & 33B & 83.1 & 82.3          & 82.8          & 76.0 & \textbf{80.0} & \textbf{57.8} & 58.6 \\
     & 65B & 85.3 & 82.8 & \textbf{84.2} & 77.0 & 78.9          & 56.0          & \textbf{60.2} \\
  \midrule
  \textbf{h2oGPT}
    & 6.9B & 61.6 & 76.8 & 67.0 & 61.6 & 65.4 & 35.6 & 38.1 \\
    & 12B  & 66.9 & 76.6 & 68.0 & 63.7 & 62.2 & 35.1 & 37.4 \\
    & 20B  & 71.3 & 77.8 & 72.6 & 66.1 & 68.9 & 44.2 & 40.0 \\
    & 40B  & 85.2 & \textbf{83.3} & 83.1 & 77.5 & 78.0 & 54.6 & 48.8 \\

  \bottomrule
  \end{tabular}
  \caption{
  \textbf{Zero-shot performance on Common Sense Reasoning tasks. Other scores from \href{https://arxiv.org/abs/2302.13971}{LLaMa paper}}.
  \label{tab:commonsense}
  }
\end{table*}

We also used \href{https://sharegpt.com/}{ShareGPT} prompts and evaluated the answers provided by h2oGPT by asking the OpenAssistant reward model or an advanced LLM like GPT-3.5/4 for a score between 0 and 1, or for which of two answers is better. More details can be found on our GitHub repositories.

\subsubsection{Current Weaknesses}
h2oGPT fine-tuned LLMs exhibit the same biases and limitations as their underlying foundation models, including:

\begin{itemize}
    \item Factual correctness
    \item Code completion
    \item Reasoning, chain-of-thought
    \item Mathematics and logic
\end{itemize}

\subsubsection{Current Capabilities}
h2oGPT fine-tuned LLMs exhibit certain capabilities that can exceed their underlying foundation models without requiring significant prompt engineering:
\begin{itemize}
\item General Chat
\item Summarization
\item Creativity
\item Rephrasing
\item Private Document Chat with fact-based answers (thanks to VectorDB integration)
\end{itemize}

\section{Outlook}
There are several roadmap items we intend to work on in the near future, but these might change based on customer/community feedback or new developments:
\begin{itemize}
    \item Reinforcement Learning with Human Feedback in H2O LLM Studio
    \item Improved VectorDB document search using metadata, large-context, prompt-to-code generation
    \item \href{https://arxiv.org/abs/2304.12244}{Wizard LM} for automatic high-quality data preparation
    \item \href{https://arxiv.org/abs/2305.03047}{Self-alignment} (research)
    \item Use the latest available open-source models and techniques for architectural or data-specific improvements
\end{itemize}


\section{Conclusion}
We are excited to announce that we have open-sourced a range of essential code components that are instrumental in effectively fine-tuning Language Models (LLMs) and transforming them into advanced ChatBots and Document Search engines. Our commitment to open-source principles means that we provide 100\% permissive access to data, models, and code, empowering the wider community to leverage and build upon our advancements.

Through our extensive research and development efforts, we have achieved the cutting-edge in data preparation and fine-tuning techniques for LLMs. The resulting models represent the state of the art in the field, while adhering to commercially viable licenses. We remain dedicated to maintaining our position at the forefront of the learning curve, continuously pushing the boundaries of what is achievable.

It's important to note that our existing products, such as \href{https://h2o.ai/platform/ai-cloud/make/h2o-driverless-ai/}{H2O Driverless AI}, \href{https://h2o.ai/platform/ai-cloud/make/hydrogen-torch/}{H2O Hydrogen Torch}, and \href{https://h2o.ai/platform/ai-cloud/make/document-ai/}{H2O Document AI}, have already incorporated LLMs and other deep learning models for several years. By harnessing the power of the GPT revolution, we ensure that all our products continue to benefit from the ongoing innovations in this rapidly evolving field.

We are excited to contribute to the advancement of the NLP community and look forward to the collective progress that will be accelerated by the availability of our open-sourced code and models.

\clearpage

\section*{References}
This is partial list of references that we collected during the creation of h2oGPT. We'd like to thank all collaborators and open-source community members.

\subsection*{h2oGPT repositories and discussion channels}
\begin{itemize}
\item \href{https://github.com/h2oai/h2ogpt}{\faGithubSquare h2oGPT \texttt{https://github.com/h2oai/h2ogpt}}
\item \href{https://github.com/h2oai/h2o-llmstudio}{\faGithubSquare H2O LLM Studio \texttt{https://github.com/h2oai/h2o-llmstudio}}
\item \href{https://huggingface.co/h2oai}{H2O.ai on Hugging Face \texttt{https://huggingface.co/h2oai}}
\item \href{https://discord.com/channels/1097462770674438174/1100717863221870643}{H2O.ai Generative Discord Channel}
\end{itemize}

\subsection*{LLM related code directly used for h2oGPT:}
\begin{itemize}
    \item \href{https://github.com/h2oai/alpaca-lora}{Alpaca LoRa}
    \item \href{https://github.com/microsoft/LoRA}{LoRa}
    \item \href{https://github.com/huggingface/transformers}{Hugging Face Transformers}
    \item \href{https://github.com/huggingface/datasets}{Hugging Face Datasets}
    \item \href{https://github.com/huggingface/peft}{Hugging Face PEFT}
    \item \href{https://github.com/TimDettmers/bitsandbytes}{bitsandbytes}
    \item \href{https://github.com/pytorch/pytorch}{PyTorch}
    \item \href{https://github.com/PanQiWei/AutoGPTQ}{AutoGPTQ}
\end{itemize}

\subsection*{Code to consider including:}
\begin{itemize}
    \item \href{https://github.com/declare-lab/flan-alpaca}{flan-alpaca}
    \item \href{https://github.com/oobabooga/text-generation-webui}{text-generation-webui}
    \item \href{https://github.com/zphang/minimal-llama/}{minimal-llama}
    \item \href{https://nn.labml.ai/neox/samples/finetune.html}{finetune GPT-NeoX}
    \item \href{https://github.com/qwopqwop200/GPTQ-for-LLaMa}{GPTQ for LLaMa}
    \item \href{https://github.com/togethercomputer/OpenChatKit/issues/20}{OpenChatKit on multi-GPU}
    \item \href{https://huggingface.co/docs/transformers/main/en/model_doc/gptj#transformers.GPTJForSequenceClassification}{Non-Causal LLM}
    \item \href{https://github.com/togethercomputer/OpenChatKit/commit/148b5745a57a6059231178c41859ecb09164c157}{OpenChatKit Offload}
    \item \href{https://github.com/declare-lab/flan-alpaca/blob/main/training.py}{Flan-alpaca}
\end{itemize}

\subsection*{Some open source models:}
\begin{itemize}
    \item \href{https://huggingface.co/togethercomputer/GPT-NeoXT-Chat-Base-20B/tree/main}{GPT-NeoXT-Chat-Base-20B}
    \item \href{https://huggingface.co/docs/transformers/model_doc/gpt_neox}{GPT-NeoX}
    \item \href{https://huggingface.co/EleutherAI/gpt-neox-20b}{GPT-NeoX-20B}
    \item \href{https://huggingface.co/EleutherAI/pythia-6.9b}{Pythia-6.9B}
    \item \href{https://huggingface.co/EleutherAI/neox-ckpt-pythia-12b}{Pythia-12B}
    \item \href{https://huggingface.co/google/flan-t5-xxl}{Flan-T5-XXL}
    \item \href{https://huggingface.co/togethercomputer/GPT-JT-Moderation-6B}{GPT-J-Moderation-6B}
    \item \href{https://laion.ai/blog/oig-dataset/#safety-models}{OIG safety models}
    \item \href{https://huggingface.co/mT0}{BigScience-mT0}
    \item \href{https://huggingface.co/datasets/bigscience/xP3}{BigScience-XP3}
    \item \href{https://huggingface.co/bigscience/bloomz}{BigScience-Bloomz}
\end{itemize}

\subsection*{Some creative commons models that would be interesting to use:}
\begin{itemize}
    \item \href{https://huggingface.co/facebook/galactica-120b}{Galactica-120B}
    \item \href{https://huggingface.co/decapoda-research/llama-smallint-pt}{LLaMa-small-pt}
\item \href{https://huggingface.co/maderix/llama-65b-4bit/tree/main}{LLaMa-64b-4bit}
\end{itemize}

\subsection*{Papers/Repos}
\begin{itemize}
    \item \href{https://arxiv.org/abs/2210.11610}{Self-improve}
    \item \href{https://arxiv.org/abs/2303.17491}{Coding}
    \item \href{https://arxiv.org/abs/2303.11366}{self-reflection}
    \item \href{https://arxiv.org/abs/2204.05862}{RLHF}
    \item \href{https://arxiv.org/abs/2303.17071}{DERA}
    \item \href{https://aiindex.stanford.edu/report/}{HAI Index Report 2023}
    \item \href{https://arxiv.org/abs/2302.13971}{LLaMa}
    \item \href{https://github.com/THUDM/GLM-130B}{GLM-130B}
    \item \href{https://github.com/BlinkDL/RWKV-LM}{RWKV RNN}
    \item \href{https://arxiv.org/abs/2302.04761}{Toolformer}
    \item \href{https://github.com/qwopqwop200/GPTQ-for-LLaMa}{GPTQ}
    \item \href{https://www.deepmind.com/publications/improving-language-models-by-retrieving-from-trillions-of-tokens}{Retro}
    \item \href{https://arxiv.org/abs/2302.08091}{Clinical outperforms}
    \item \href{https://github.com/amazon-science/mm-cot}{Chain-Of-Thought}
    \item \href{https://arxiv.org/abs/2203.15556}{scaling law1}
    \item \href{https://github.com/google/BIG-bench}{Big-bench}
    \item \href{https://github.com/allenai/natural-instructions}{Natural-Instructions}
\end{itemize}

\subsection*{Other projects:}
\begin{itemize}
    \item \href{https://huggingface.co/blog/stackllama}{StackLLaMa}
    \item \href{https://github.com/PhoebusSi/alpaca-CoT}{Alpaca-CoT}
    \item \href{https://github.com/hpcaitech/ColossalAI/tree/main/applications/Chat}{ColossalAIChat}
    \item \href{https://github.com/young-geng/EasyLM.git}{EasyLM}
    \item \href{https://bair.berkeley.edu/blog/2023/04/03/koala/}{Koala}
    \item \href{https://vicuna.lmsys.org/}{Vicuna}
    \item \href{https://github.com/declare-lab/flan-alpaca}{Flan-Alpaca}
    \item \href{https://chat.lmsys.org/}{FastChat}
    \item \href{https://github.com/Nuked88/alpaca.http}{alpaca.http}
    \item \href{https://github.com/openai/chatgpt-retrieval-plugin}{chatgpt-retrieval-plugin}
    \item \href{https://www.subtl.ai/}{subtl.ai docs search on private docs}
    \item \href{https://gretel.ai/}{gretel}
    \item \href{https://github.com/johnsmith0031/alpaca_lora_4bit}{alpaca lora 4bit}
    \item \href{https://github.com/s4rduk4r/alpaca_lora_4bit_readme}{alpaca lora 4bit readme}
    \item \href{https://github.com/sahil280114/codealpaca}{code alpaca}
    \item \href{https://github.com/nsarrazin/serge}{serge}
    \item \href{https://github.com/BlinkDL}{BlinkDL}
    \item \href{https://github.com/mosaicml/examples#large-language-models-llms}{MosaicCM}
    \item \href{https://openai.com/blog/chatgpt-plugins}{OpenAI Plugins}
    \item \href{https://github.com/gannonh/gpt3.5-turbo-pgvector}{GPT3.5-Turbo-PGVector}
    \item \href{https://github.com/ZrrSkywalker/LLaMA-Adapter}{LLaMa-Adapter}
    \item \href{https://github.com/jerryjliu/llama_index}{llama-index}
    \item \href{https://github.com/zphang/minimal-llama/}{minimal-llama}
    \item \href{https://github.com/ggerganov/llama.cpp}{llama.cpp}
    \item \href{https://justine.lol/mmap/}{mmap}
    \item \href{https://til.simonwillison.net/llms/llama-7b-m2}{lamma.cpp more}
    \item \href{https://github.com/helliun/targetedSummarization}{TargetedSummarization}
    \item \href{https://laion.ai/blog/open-flamingo/}{OpenFlamingo}
    \item \href{https://github.com/Torantulino/Auto-GPT}{Auto-GPT}
    \item \href{https://github.com/imartinez/privateGPT}{PrivateGPT}
\end{itemize}

\subsection*{Apache2/etc. Data}
\begin{itemize}
    \item \href{https://laion.ai/blog/oig-dataset/}{OIG 43M instructions} (\href{https://huggingface.co/datasets/laion/OIG}{direct HF link})
    \item \href{https://laion.ai/blog/oig-dataset/}{More on OIG}
    \item \href{https://huggingface.co/datasets/viewer/?dataset=squad}{DataSet Viewer}
    \item \href{https://huggingface.co/datasets/Anthropic/hh-rlhf}{Anthropic RLHF}
    \item \href{https://huggingface.co/datasets/openai/webgpt_comparisons}{WebGPT\_Comparisons}
    \item \href{https://github.com/yizhongw/self_instruct}{Self\_instruct}
    \item \href{https://github.com/togethercomputer/OpenDataHub}{20BChatModelData}
\end{itemize}

\subsection*{Apache2/MIT/BSD-3 Summarization Data}
\begin{itemize}
    \item \href{https://huggingface.co/datasets/xsum}{xsum for Summarization}
    \item \href{https://huggingface.co/datasets?task_categories=task_categories:summarization&license=license:apache-2.0&sort=downloads}{Apache2 Summarization}
    \item \href{https://huggingface.co/datasets?task_categories=task_categories:summarization&license=license:mit&sort=downloads}{MIT summarization}
    \item \href{https://huggingface.co/datasets?task_categories=task_categories:summarization&license=license:bsd-3-clause&sort=downloads}{BSD-3 summarization}
    \item \href{https://huggingface.co/datasets?task_categories=task_categories:summarization&license=license:openrail&sort=downloads}{OpenRail}
    \item \href{https://huggingface.co/datasets/openai/summarize_from_feedback}{Summarize\_from\_feedback}
\end{itemize}

\subsection*{Ambiguous License Data}
\begin{itemize}
    \item \href{https://github.com/Instruction-Tuning-with-GPT-4/GPT-4-LLM}{GPT-4-LLM}
    \item \href{https://huggingface.co/datasets/nomic-ai/gpt4all_prompt_generations}{GPT4All}
    \item \href{https://github.com/lm-sys/FastChat/issues/90#issuecomment-1493250773}{LinkGPT4}
    \item \href{https://huggingface.co/datasets/RyokoAI/ShareGPT52K}{ShareGPT52K}
    \item \href{https://huggingface.co/datasets/anon8231489123/ShareGPT_Vicuna_unfiltered}{ShareGPT\_Vicuna}
    \item \href{https://chatlogs.net/}{ChatLogs}
    \item \href{https://github.com/PhoebusSi/alpaca-CoT}{Alpaca-CoT}
    \item \href{https://github.com/mbzuai-nlp/LaMini-LM}{LaMini-LM}
\end{itemize}

\subsection*{Non-commercial Data}
\begin{itemize}
    \item \href{https://github.com/gururise/AlpacaDataCleaned}{GPT-3 based Alpaca Cleaned}
    \item \href{https://github.com/databrickslabs/dolly/tree/master}{Dolly}
\end{itemize}

\subsection*{Prompt Engineering}
\begin{itemize}
    \item \href{https://github.com/huggingface/peft}{PEFT Prompt/P-tuning}
    \item \href{https://docs.nvidia.com/deeplearning/nemo/user-guide/docs/en/main/nlp/nemo_megatron/prompt_learning.html}{Prompt/P-tuning Nemo/NVIDIA}
    \item \href{https://lilianweng.github.io/posts/2023-03-15-prompt-engineering/}{Info}
    \item \href{https://github.com/dair-ai/Prompt-Engineering-Guide}{Info2}
    \item \href{https://arxiv.org/abs/2104.08691}{Prompt-Tuning}
    \item \href{https://arxiv.org/abs/2110.07602}{P-tuning v2}
    \item \href{https://github.com/yoheinakajima/babyagi/blob/main/babyagi.py#L97-L134}{babyagi}
\end{itemize}

\subsection*{Validation}
\begin{itemize}
    \item \href{https://arize.com/blog-course/generative-ai-metrics-bleu-score/}{Bleu/Rouge/Meteor/Bert-Score}
    \item \href{https://github.com/EleutherAI/lm-evaluation-harness}{LM Evaluation Harness}
\end{itemize}

\subsection*{Generate Hyperparameters}
\begin{itemize}
    \item \href{https://huggingface.co/blog/how-to-generate}{hot-to-generate}
    \item \href{https://christianjmills.com/posts/transformers-book-notes/chapter-5/index.html}{Notes\_on\_Transformers Chpt5}
    \item \href{https://christianjmills.com/posts/transformers-book-notes/chapter-10/index.html}{Notes\_on\_Transformers\_Chpt10}
\end{itemize}

\subsection*{Embeddings}
\begin{itemize}
    \item \href{https://medium.com/@nils_reimers/openai-gpt-3-text-embeddings-really-a-new-state-of-the-art-in-dense-text-embeddings-6571fe3ec9d9}{OpenAI Expensive?}
    \item \href{https://huggingface.co/spaces/mteb/leaderboard}{Leaderboard}
\end{itemize}

\subsection*{Commercial products}
\begin{itemize}
    \item \href{https://platform.openai.com/docs/guides/fine-tuning/advanced-usage}{OpenAI}
    \item \href{https://platform.openai.com/tokenizer}{OpenAI Tokenizer}
    \item \href{https://platform.openai.com/playground}{OpenAI Playground}
    \item \href{https://chat.openai.com/chat?}{OpenAI Chat}
    \item \href{https://chat.openai.com/chat?model=gpt-4}{OpenAI GPT-4 Chat}
    \item \href{https://cohere.io/}{cohere}
    \item \href{https://docs.cohere.ai/reference/finetune}{coherefinetune}
    \item \href{https://docsbot.ai/}{DocsBotAI}
    \item \href{https://www.perplexity.ai/}{Perplexity}
    \item \href{https://www.voiceflow.com/}{VoiceFlow}
    \item \href{https://nlpcloud.com/effectively-using-gpt-j-gpt-neo-gpt-3-alternatives-few-shot-learning.html}{NLPCloud}
\end{itemize}

\subsection*{Inference}
\begin{itemize}
    \item \href{https://github.com/triton-inference-server/fastertransformer_backend#multi-node-inference}{FasterTransformer}
    \item \href{https://developer.nvidia.com/blog/deploying-nvidia-triton-at-scale-with-mig-and-kubernetes/}{Kubernetes Triton}
    \item \href{https://github.com/huggingface/optimum}{Optimum}
    \item \href{https://github.com/mlc-ai/mlc-llm}{MLC-LLM}
    \item \href{https://github.com/triton-inference-server}{Triton Inference server}
\end{itemize}

\subsection*{Semi-Open source Semi-Commercial products}
\begin{itemize}
    \item \href{https://open-assistant.io/}{OpenAssistant}
    \item \href{https://github.com/LAION-AI/Open-Assistant}{OpenAssistant Repo}
    \item \href{https://github.com/togethercomputer/OpenChatKit}{OpenChatKit}
    \item \href{https://github.com/togethercomputer/OpenDataHub}{OpenDataHub}
    \item \href{https://www.together.xyz/blog/openchatkit}{OpenChatKit3}
    \item \href{https://github.com/togethercomputer/OpenChatKit/blob/main/training/README.md#arguments}{OpenChatKit4}
    \item \href{https://python.langchain.com/en/latest/}{langchain}
    \item \href{https://www.youtube.com/watch?v=nMniwlGyX-c}{langchain+pinecone}
\end{itemize}

\subsection*{Q/A docs}
\begin{itemize}
    \item \href{https://www.humata.ai/}{HUMATA}
    \item \href{https://osschat.io/}{OSSCHat}
    \item \href{https://txt.cohere.com/embedding-archives-wikipedia/}{NeuralSearchCohere}
    \item \href{https://github.com/bublint/ue5-llama-lora}{ue5}
\end{itemize}

\subsection*{AutoGPT type projects}
\begin{itemize}
    \item \href{https://github.com/reworkd/AgentGPT}{AgentGPT}
    \item \href{https://arxiv.org/abs/2304.05128}{Self-DEBUG}
    \item \href{https://github.com/yoheinakajima/babyagi/}{BabyAGI}
    \item \href{https://github.com/irgolic/AutoPR}{AutoPR}
\end{itemize}

\subsection*{Cloud fine-tune}
\begin{itemize}
    \item \href{https://docs.aws.amazon.com/sagemaker/latest/dg/jumpstart-fine-tune.html}{AWS}
    \item \href{https://aws.amazon.com/blogs/machine-learning/training-large-language-models-on-amazon-sagemaker-best-practices/}{AWS2}
\end{itemize}

\subsection*{Chatbots}
\begin{itemize}
    \item \href{https://github.com/nomic-ai/gpt4all-chat}{GPT4ALL Chat}
    \item \href{https://github.com/nomic-ai/gpt4all}{GLT4ALL}
    \item \href{https://open-assistant.io/chat}{OASSST}
    \item \href{https://github.com/lm-sys/FastChat}{FastChat}
    \item \href{https://huggingface.co/spaces/HuggingFaceH4/databricks-dolly}{Dolly}
    \item \href{https://huggingface.co/spaces/HuggingFaceH4/instruction-model-outputs-filtered}{HF Instructions}
    \item \href{https://github.com/microsoft/DeepSpeedExamples/tree/master/applications/DeepSpeed-Chat}{DeepSpeed Chat}
    \item \href{https://github.com/bupticybee/FastLoRAChat}{LoraChat}
    \item \href{https://github.com/TabbyML/tabby}{Tabby}
    \item \href{https://github.com/dylan-slack/TalkToModel}{TalkToModel}
\end{itemize}

\subsection*{LangChain related}
\begin{itemize}
    \item \href{https://github.com/freddyaboulton/gradio-tools}{Gradio Tools}
    \item \href{https://blog.langchain.dev/gradio-llm-agents/}{LLM Agents}
    \item \href{https://github.com/mbchang/meta-prompt}{Meta Prompt}
\end{itemize}

\subsection*{Summaries}
\begin{itemize}
    \item \href{https://github.com/Mooler0410/LLMsPracticalGuide}{LLMs}
\end{itemize}

\subsection*{Hallucinations}
\begin{itemize}
    \item \href{https://dl.acm.org/doi/10.1145/3442188.3445922}{On the Dangers of Stochastic Parrots}
\end{itemize}

\section{Disclaimer}
Please read this disclaimer carefully before using the large language model provided by h2oGPT. Your use of the model signifies your agreement to the following terms and conditions.

\textbf{Biases and Offensiveness:} The large language model is trained on a diverse range of internet text data, which may contain biased, racist, offensive, or otherwise inappropriate content. By using this model, you acknowledge and accept that the generated content may sometimes exhibit biases or produce content that is offensive or inappropriate. The developers of this repository do not endorse, support, or promote any such content or viewpoints.

\textbf{Limitations:} The large language model is an AI-based tool and not a human. It may produce incorrect, nonsensical, or irrelevant responses. It is the user's responsibility to critically evaluate the generated content and use it at their discretion.

\textbf{Use at Your Own Risk:} Users of this large language model must assume full responsibility for any consequences that may arise from their use of the tool. The developers and contributors of this repository shall not be held liable for any damages, losses, or harm resulting from the use or misuse of the provided model.

\textbf{Ethical Considerations:} Users are encouraged to use the large language model responsibly and ethically. By using this model, you agree not to use it for purposes that promote hate speech, discrimination, harassment, or any form of illegal or harmful activities.

\textbf{Reporting Issues:} If you encounter any biased, offensive, or otherwise inappropriate content generated by the large language model, please report it to the repository maintainers through the provided channels. Your feedback will help improve the model and mitigate potential issues.

\textbf{Changes to this Disclaimer:} The developers of this repository reserve the right to modify or update this disclaimer at any time without prior notice. It is the user's responsibility to periodically review the disclaimer to stay informed about any changes.

By using the large language model provided in this repository, you agree to accept and comply with the terms and conditions outlined in this disclaimer. If you do not agree with any part of this disclaimer, you should refrain from using the model and any content generated by it.

Online version: \href{https://github.com/h2oai/h2ogpt#disclaimer}{Disclaimer}

\bibliographystyle{unsrt}  

\end{document}